# Large-scale Dynamic Network Representation via Tensor Ring Decomposition

Qu Wang

*Abstract*—Large-scale Dynamic Networks (LDNs) are becoming increasingly important in the Internet age, yet the dynamic nature of these networks captures the evolution of the network structure and how edge weights change over time, posing unique challenges for data analysis and modeling. A Latent Factorization of Tensors (LFT) model facilitates efficient representation learning for a LDN. But the existing LFT models are almost based on Canonical Polyadic Factorization (CPF). Therefore, this work proposes a model based on Tensor Ring (TR) decomposition for efficient representation learning for a LDN. Specifically, we incorporate the principle of single latent factor-dependent, non-negative, and multiplicative update (SLF-NMU) into the TR decomposition model, and analyze the particular bias form of TR decomposition. Experimental studies on two real LDNs demonstrate that the propose method achieves higher accuracy than existing models.

## I. Introduction

A large-scale dynamic network (LDN) is a network model that can capture the evolution of network structure and changes in edge weights over time [1]-[2]. In the Internet age, such networks have become increasingly important as they facilitate the understanding and analysis of complex data in many applications such as social networks, bioinformatics, and financial analysis. They can be represented by a LDN, where each node represents an involved node, each link represents a temporally observed interaction between two nodes, and each weight quantifies the strength of the specified interaction. Such a LDN contains rich knowledge about the behavior patterns of the involved nodes. However, the dynamic nature of these network poses unique challenges in developing effective methods to perform representation learning [3]-[5]. As a result, the study of representation learning method of large-scale dynamic networks has become an important research task in many fields.

In a LDN, since the network structure and edge weights change over time, and it is impossible to establish a full connection between nodes, the network data will usually contain a large number of missing values. The structure of a LDN is extremely high dimensional and incomplete (HDI), and how to model and analyze such highly incomplete data has become an important and challenging task [6]-[9].

According to previous research [10]-[11], it is very efficient to describe a LDN as an HDI tensor. A Latent Factorization of Tensors (LFT)-based model enable efficient representation learning for HDI tensors [12]-[17]. The LFT model can represent the spatial and temporal structures nature of an HDI tensor, which can effectively characterize LDN's dynamic patterns. However, a LFT model cannot describe non-negative data, which weakens its representation learning ability when dealing with dynamic data defined on non-negative domains, such as LDNs data. To address this issue, Non-negative Latent Factorization of Tensor (NLFT)-based models are proposed [18]-[21]. But existing NLFT-based models mostly use Canonical Polyadic Factorization (CPF) [22]-[28]. Although a NLFT model based on CPT is computationally efficient, it may ignore complex structures in the data [29]-[31]. In contrast, Tensor-Ring decomposition can maintain low-rank properties while preserving structural information, and are relatively computationally efficient for high-order tensors [32]-[33]. In this study, we adopt TR decomposition to build a NLFT model, which can capture more features in an HDI tensor describing a LDN. Furthermore, this study analyzes the bias form specific to the TR decomposition model and incorporate it into the learned model to describe the data fluctuations.

In summary, the main contributions of this study are as follows:

1) We propose a NLFT model based on TR decomposition to efficiently represent a LDN, which use the principle of SLF-UMN to greatly reduce the complexity of the proposed model.

2) We introduce linear bias (LB) into this model to enhance the representation learning capability of the proposed model.

The validation experiments are conducted on two real LDNs datasets, and the experimental results show that the proposed model outperforms the existing models in terms of prediction accuracy.

## II. Preliminaries

### A. Tensor Ring Decomposition

TR decomposition is first proposed in [34]. Due to its powerful expressive ability, it has received a lot of attention in recent years. Due to limited space, we briefly introduce the definition of TR decomposition and the formulas used.

**Definition 1:** (Tensor Ring decomposition): TR decomposition refers to decomposing a high-order tensor into a sequence of third-order tensors that are cyclically multiplied [35]-[36]. Specifically, for an $N$th-order tensor $\mathbf{X}^{I_1 \times \cdots \times I_N}$, which can be represented by a series of third-order small-large tensors $\mathbf{Z}^{(k)} \in \mathbb{R}^{r_{k-1} \times I_k \times r_k}, k=1,2,\ldots,N$, where $r_0 = r_N$. The definition of TR decomposition is given below.

$$\mathbf{X} = R\left(\mathbf{Z}^{(1)},\ldots,\mathbf{Z}^{(N)}\right) \qquad (1)$$

where R(.) represents the TR composition operation on tensors, and $\mathbf{X}$'s each element can be formulated as:

---

$$\mathbf{X}(i_1, i_2, \ldots, i_N) = \mathrm{tr}\left(Z_{i_1}^{(1)}, Z_{i_2}^{(2)}, \ldots, Z_{i_N}^{(N)}\right) \tag{2}$$

where $Z_{i_k}^{(k)} \in \mathbb{R}^{r_{k-1} \times r_k}$ represents the $i_k$th lateral slice matrix of $\mathbf{Z}^{(k)}$, and tr(.) is the operation to find the matrix trace. In addition, the above formula can also be expressed by using the product of individual elements.

$$\mathbf{X}(i_1, i_2, \ldots, i_N) = \sum_{r_1=1}^{R_1} \sum_{r_2=1}^{R_2} \ldots \sum_{r_N=1}^{R_N} z_{r_0 i_1 r_1}^{(1)} \times z_{r_1 i_2 r_2}^{(2)} \times \ldots \times z_{r_{N-1} i_N r_N}^{(N)} \tag{3}$$

The $z_{r_{k-1} i_k r_k}^{(k)}$ denotes the element at position $(r_{k-1}, i_k, r_k)$ of the tensor $\mathbf{Z}^{(k)}$, $k = 1, 2, \ldots, N$. The set $\{R_1, R_2, \ldots, R_N\}$ are called TR-ranks, $1 \leq r_k \leq R_k$.

### B. A LDN Data Tensor

***Definition 2:*** (An HDI tensor for a LDN): Given an HDI tensor $\mathbf{Y}^{|I| \times |J| \times |K|}$ describing a LDN, an element $y_{ijk}$ denotes the weight of a directed edge from node $i \in I$ to node $j \in J$ at time point $k \in K$. An HDI tensor means that the known elements set $\Lambda$ is much smaller than the unknown elements set $\Gamma$, which is symbolized as $|\Lambda| \ll |\Gamma|$.

### C. Problem Formulation

In this work, we apply TR decomposition to performs the latent factorization of tensors on $\mathbf{Y}$. We decompose the tensor $\mathbf{Y}$ into three low-rank core tensors $\mathbf{U}^{R_3 \times |I| \times R_1}, \mathbf{V}^{R_1 \times |J| \times R_2}, \mathbf{W}^{R_2 \times |K| \times R_3}$, which is a low-rank approximation of the tensor $\mathbf{Y}$.

$$\hat{\mathbf{Y}} = R(\mathbf{U}, \mathbf{V}, \mathbf{W}) \tag{4}$$

According to (2) and (3), we can get the expression of a single element of $\hat{\mathbf{Y}}$. Note that for the convenience of calculation and analysis, we set the rank $R$ of Tensor-Ring as $R = R_1 = R_2 = \ldots = R_N$.

$$\hat{y}_{ijk} = \mathrm{Tr}(U_i V_j W_k) = \sum_{r_1=1}^{R} \sum_{r_2=1}^{R} \sum_{r_3=1}^{R} u_{r_3 i r_1} v_{r_1 j r_2} w_{r_2 k r_3} \tag{5}$$

To obtain the desired LF tensor $\mathbf{U}, \mathbf{V}, \mathbf{W}$, we build an objective function to measure the difference between $\mathbf{Y}$ and $\hat{\mathbf{Y}}$ in LFT [37]-[38]. In this paper, we use the commonly used Euclidean distance to construct measure function. Since $\mathbf{Y}$ is a HDI tensor, we only need to model this difference based on the set of known elements $\Lambda$ [39]-[40], according to the above principles, the objective function is defined as:

$$\varepsilon = \sum_{y_{ijk} \in \Lambda} \left( y_{ijk} - \sum_{r_1=1}^{R_1} \sum_{r_2=1}^{R_2} \sum_{r_3=1}^{R_3} u_{r_3 i r_1} v_{r_1 j r_2} w_{r_2 k r_3} \right)^2 \tag{6}$$

In order to obtain the optimal solution of the ill-posed non-convex objective function and improve the generality of the model, the $L_2$-norm-based regularization is incorporated into the above objective function. Note that the data in a HDI tensor describing a LDN are all non-negative. In order to model such non-negative data more accurately, the elements in $\mathbf{U}, \mathbf{V}, \mathbf{W}$ need to be constrained to be non-negative. Then (6) can be redefined as follows

$$\varepsilon = \sum_{y_{ijk} \in \Lambda} \left( \left( y_{ijk} - \sum_{r_1=1}^{R} \sum_{r_2=1}^{R} \sum_{r_3=1}^{R} u_{r_3 i r_1} v_{r_1 j r_2} w_{r_2 k r_3} \right)^2 + \lambda_1 \left( \sum_{r_1=1}^{R} \sum_{r_3=1}^{R} u_{r_3 i r_1}^2 + \sum_{r_1=1}^{R} \sum_{r_2=1}^{R} v_{r_1 j r_2}^2 + \sum_{r_2=1}^{R} \sum_{r_3=1}^{R} w_{r_2 k r_3}^2 \right) \right) \tag{7}$$

$$s.t. \ \forall i \in I, \forall j \in J, \forall k \in K, r_1, r_2, r_3 \in \{1, \ldots, R\} : u_{r_3 i r_1} \geq 0, v_{r_1 j r_2} \geq 0, w_{r_2 k r_3} \geq 0.$$

where $\lambda_1$ denote the regularization coefficients for LF tensor.

## III. OUR MODEL

### A. Linear Bias Modeling in TR Decomposition

In a real application, the network structure of a LDN changes over time, accordingly, link weight also changes. In order to represent LDN better and efficiently, it is very important to introduce linear bias into the learning model for modeling the fluctuation of HDI tensor data. Next, we introduce a bias tensor $\mathbf{L}$ into the problem, and it should be the same size as $\hat{\mathbf{Y}}$. Similarly, we use TR decomposition to decompose the tensor $\mathbf{L}$ into three low-rank core tensors $\mathbf{A}^{R \times |I| \times R}, \mathbf{B}^{R \times |J| \times R}, \mathbf{C}^{R \times |K| \times R}$.

$$\mathbf{L} = R(\mathbf{A}, \mathbf{B}, \mathbf{C}) \tag{8}$$

As shown in Figure 1, linear bias in LFT rely on three bias tensors, which correspond to the three dimensions of $\hat{\mathbf{Y}}$ respectively. For each tensor, only diagonal slices have valid elements, and the remaining elements are filled with the constant 0. Each element of the bias tensor $\mathbf{L}$ can be obtained by the trace of the matrix multiplication of the corresponding linear bias tensor lateral slices.

$$l_{ijk} = \mathrm{Tr}(A_i B_j C_k) \tag{9}$$

Where $A_i, B_j, C_k$ represents the $i$th, $j$th, $k$th lateral slice matrix of linear bias tensor $\mathbf{A}, \mathbf{B}, \mathbf{C}$, respectively. In order to simplify the calculation, we use matrix D, E, F to represent the diagonal slices of the tensor $\mathbf{A}, \mathbf{B}, \mathbf{C}$, respectively. We find that the matrix $A_i, B_j, C_k$ only has elements on the diagonal, and the corresponding diagonals are represented by the vector $d_i, e_j, f_k$, that $d_i = \text{diag}(A_i), e_j = \text{diag}(B_j), f_k = \text{diag}(C_k)$, $d_i, e_j, f_k$ is the $i$th, $j$th, $k$th row vector of D, E, F. Then, the trace of the matrix product can be represented by the inner product of the corresponding diagonal vectors. Thus, (9) is reformulated into

$$l_{ijk} = \langle d_i, e_j, f_k \rangle = \sum_{r=1}^{R} d_{ir} e_{jr} f_{kr} \tag{10}$$

By substituting (10) into (7), we get a biased objective function:

$$\varepsilon = \sum_{y_{ijk} \in \Lambda} \left( \left( y_{ijk} - \sum_{r_1=1}^{R} \sum_{r_2=1}^{R} \sum_{r_3=1}^{R} u_{r_3 i r_1} v_{r_1 j r_2} w_{r_2 k r_3} - \sum_{r=1}^{R} d_{ir} e_{jr} f_{kr} \right)^2 \right. \tag{11}$$
$$\left. + \lambda_1 \left( \sum_{r_1=1}^{R} \sum_{r_3=1}^{R} u_{r_3 i r_1}^2 + \sum_{r_1=1}^{R} \sum_{r_2=1}^{R} v_{r_1 j r_2}^2 + \sum_{r_2=1}^{R} \sum_{r_3=1}^{R} w_{r_2 k r_3}^2 \right) + \lambda_2 \sum_{r=1}^{R} \left( d_{ir}^2 + e_{jr}^2 + f_{kr}^2 \right) \right)$$

$s.t.\ \forall i \in I, \forall j \in J, \forall k \in K, r_1, r_2, r_3, r \in \{1, \ldots, R\}$:
$u_{r_3 i r_1} \geq 0, v_{r_1 j r_2} \geq 0, w_{r_2 k r_3} \geq 0,\ d_{ir} \geq 0, e_{jr} \geq 0, f_{kr} \geq 0.$

Similarly, we also add regularization terms to LBs, where $\lambda_2$ denotes the regularization coefficient for linear bias matrices.

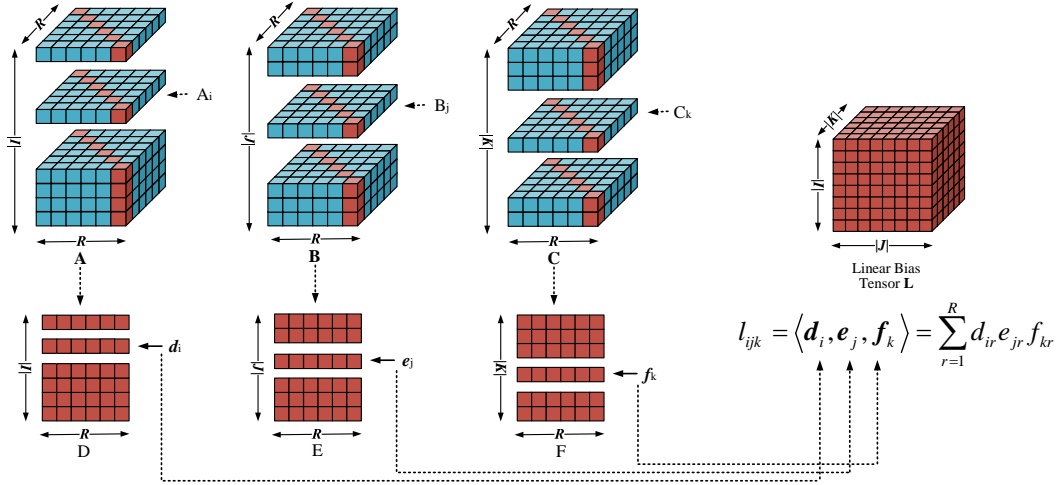

Fig. 1. Linear bias matrices and linear tensor

*B. Parameter Learning Principles*

As mentioned above, the HDI tensor data is defined in the non-negative real number field. In order to better describe this non-negative property, we constrain the parameters of the proposed model to be non-negative. According to previous research [44]-[46], the principle of single latent factor-dependent, non-negative, and multiplicative update (SLF-NMU) is usually adopt to build a NLFT model based on Canonical Polyadic Factorization. Consequently, in this study, we also utilize the SLF-NMU algorithm to build a tensor ring-based NLFT model. We first perform additive gradient descent (AGD) on each LF tensor and linear bias in (11), note that since the update rules of **U**, **V**, **W** are similar and the update rules of D, E, F are also similar, we only present **U** and D's update rules as follows:

$$\arg\min_{\mathbf{U}, d} \varepsilon \overset{AGD}{\Rightarrow} \forall i \in I, \forall j \in J, \forall k \in K, r_1, r_2, r_3, r \in \{1, 2, \ldots, R\}$$

$$\begin{cases} u_{r_3 i r_1} \leftarrow u_{r_3 i r_1} - \eta_{r_3 i r_1} \sum_{y_{ijk} \in \Lambda(i)} \left( (y_{ijk} - \hat{y}_{ijk}) \left( -\sum_{r_2=1}^{R} v_{r_1 j r_2} w_{r_2 k r_3} \right) \right) + \lambda_1 u_{r_3 i r_1} \\ d_{ir} \leftarrow d_{ir} - \eta_{ir} \left( \sum_{y_{ijk} \in \Lambda(i)} \left( (y_{ijk} - \hat{y}_{ijk}) (-e_{jr} f_{kr}) + \lambda_2 d_{ir} \right) \right) \end{cases} \tag{12}$$

where $\eta_{r_3 i r_1}$ and $\eta_{ir}$ denote the learning rates of $u_{r_3 i r_1}$ and $d_{ir}$, and $\Lambda_{(i)}$ is the subsets of $\Lambda$ associated with node $i \in V$. Note the negative terms appear in (12). We can set $\eta_{r_3 i r_1}$ and $\eta_{ir}$ respectively to ensure the nonnegativity of $u_{r_3 i r_1}$ and $d_{ir}$ during the update process.

$$\begin{cases} \eta_{r_3 i r_1} = u_{r_3 i r_1} \Big/ \sum_{y_{ijk} \in \Lambda(i)} \left( \hat{y}_{ijk} \sum_{r_2=1}^{R} v_{r_1 j r_2} w_{r_2 k r_3} + \lambda_1 u_{r_3 i r_1} \right) \\ \eta_{ir} = d_{ir} \Big/ \sum_{y_{ijk} \in \Lambda(i)} \left( \hat{y}_{ijk} e_{jr} f_{kr} + \lambda_2 d_{ir} \right) \end{cases} \tag{13}$$

Substituting (13) into (12), we can get a learning scheme based on the principle of SLF-NMU.

$$\underset{\mathbf{U},d}{\arg\min}\ \varepsilon \overset{SLL-NMU}{\Rightarrow} \forall i \in I, \forall j \in J, \forall k \in K, r_1, r_2, r_3, r \in \{1,2,\ldots,R\}$$

$$\begin{cases} u_{r_3 i r_1} \leftarrow u_{r_3 i r_1} \sum_{y_{ijk} \in \Lambda(i)} \left( y_{ijk} \sum_{r_2=1}^{R} v_{r_1 j r_2} w_{r_2 k r_3} \right) \Big/ \sum_{y_{ijk} \in \Lambda(i)} \left( \hat{y}_{ijk} \sum_{r_2=1}^{R} v_{r_1 j r_2} w_{r_2 k r_3} + \lambda_1 u_{r_3 i r_1} \right) \\ d_{ir} \leftarrow d_{ir} \left( \sum_{y_{ijk} \in \Lambda(i)} y_{ijk} e_{jr} f_{kr} \Big/ \sum_{y_{ijk} \in \Lambda(i)} \left( \hat{y}_{ijk} e_{jr} f_{kr} + \lambda_2 d_{ir} \right) \right) \end{cases} \quad (14)$$

## IV. EXPERIMENTAL RESULTS AND DISCUSSION

### A. General Settings

*1) Datasets:* We conduct experiments on two real LDN datasets, which are collected from a real terminal interaction pattern analysis system, involving personal terminal interaction data flow in a certain city in China. Note that for beneficial needs, all node ID are encrypted and all interaction data are converted to weights in the scale of (0,10). According to the above principles, four LDNs are established, the details of which are shown in Table I.

For each data set, we randomly divide the known elements into three subsets with ratio 7:1:2, corresponding to the training set $\Phi$ used to train the model, the validating set $\Psi$ to judge the model training situation, and the testing set $\Omega$ to evaluate the model performance.

TABLE I. Dataset Details

| Datasets | Nodes | Time Points | Entries | Density |
|---|---|---|---|---|
| **D1** | 409028 | 1149 | 3122420 | $1.62\times10^{-8}$ |
| **D2** | 298280 | 865 | 2055422 | $2.67\times10^{-8}$ |

*2) Evaluation Metrics:* In this paper, we choose RMSE and MAE to reflect the prediction accuracy of the model on the missing data of the HDI tensor. Lower RMSE and MAE indicate higher prediction accuracy of the model for tensor missing data.

$$\text{RMSE} = \sqrt{\frac{\sum_{y_{ijk} \in \Omega} \left( y_{ijk} - \hat{y}_{ijk} \right)^2}{|\Omega|}},$$

$$\text{MAE} = \frac{1}{|\Omega|} \sum_{y_{ijk} \in \Omega} \left| y_{ijk} - \hat{y}_{ijk} \right|.$$

### B. Compared Models

In the next, we compare our model with the following four advanced models.

M1: A LFT model based on CP decomposition [47], which adopts an alternating least squares as optimization algorithms.
M2: A multidimensional data forecasting method [48], which constructs multi-linear algebra on a HDI tensor.
M3: A biased NLFT model [20], which utilizes the CP decomposition to build NLFT model.
M4: The model proposed in this paper.

### C. Experimental Results

The experimental results of M1-4 are depicted in Table II respectively. From these results, we can see that the prediction accuracy of our proposed model M4 is higher than its peers. For instance, M4 achieves its lowest RMSE at 0.2775 on D1, while other three comparison models achieve 0.3174, 0.3496, and 0.2862 respectively. Hence, M4 has 12.57%, 20.62%, and 3.03% accuracy gain over M1, M2, M3 in RMSE on D1. On the other hand, on D1, the lowest MAE obtained by benchmarks is 0.1734, 0.1999, 0.2074, and 0.1946, respectively. Compared with three comparison models, M4 has 13.25%, 16.39%, and 10.89% accuracy gain over M1, M2, M3 in MAE on D1. On D2, M4's RMSE is 0.2812, which is about 16.10% lower than M1's 0.3352, 24.63% lower than M2's 0.3731, and 3.79% lower than M3's 0.2923. Considering MAE, the outputs by M1-4 are 0.2132, 0.2146, 0.2013, and 0.1799, respectively. Thus, M4 also achieves much lower MAE than its peers do.

TABLE II. Lowest RMSE and MAE of Each Model on All Testing Case.

| Datasets | | M1 | M2 | M3 | M4 |
|---|---|---|---|---|---|
| **D1** | RMSE | 0.3174 | 0.3496 | 0.2862 | **0.2775** |
| | MAE | 0.1999 | 0.2074 | 0.1946 | **0.1734** |
| **D2** | RMSE | 0.3352 | 0.3731 | 0.2923 | **0.2812** |
| | MAE | 0.2132 | 0.2146 | 0.2013 | **0.1799** |

## V. Conclusion

In this study, we propose an NLFT model based on the TR decomposition framework, and the SLF-NMU algorithm is applied to build the proposed model. In particular, we introduce linear bias to model the data fluctuation of LDNs, and analyze the form of linear bias under the TR decomposition framework. Finally, we conduct experiments on two real LDNs, the experimental results demonstrate that our proposed model has higher prediction accuracy.

## References

[1] S. Li, M. Zhou, X. Luo and Z. You, "Distributed Winner-Take-All in Dynamic Networks," *IEEE Transactions on Automatic Control*, vol. 62, no. 2, pp. 577-589, 2017.

[2] Y. Han, G. Huang, S. Song, L. Yang, H. Wang, and Y. Wang. "Dynamic neural networks: A survey," *IEEE Transactions on Pattern Analysis and Machine Intelligence*, vol. 44, no.10, pp. 7436-7456, 2021.

[3] S. Li, M. Zhou and X. Luo, "Modified Primal-Dual Neural Networks for Motion Control of Redundant Manipulators with Dynamic Rejection of Harmonic Noises," *IEEE Transactions on Neural Networks and Learning Systems*, vol. 29, no. 10, pp. 4791-4801, 2018.

[4] J. Jiang, and Y. Lai. "Model-free prediction of spatiotemporal dynamical systems with recurrent neural networks: Role of network spectral radius," *Physical Review Research*, vol. 1, no. 3, pp. 033-056, 2019.

[5] Q. Xuan, Z. Zhang, C. Fu, H. Hu, and V. Filkov, "Social synchrony on complex networks," *IEEE Trans. on Cybernetic*, vol. 48, no. 5, pp. 1420–1431, 2018.

[6] X. Luo, H. Wu, and Z. Li, "NeuLFT: A Novel Approach to Nonlinear Canonical Polyadic Decomposition on High-Dimensional Incomplete Tensors," *IEEE Transactions on Knowledge and Data Engineering*, DOI: 10.1109/TKDE.2022.3176466.

[7] H. Wu, and X. Luo. "Instance-Frequency-Weighted Regularized, Nonnegative and Adaptive Latent Factorization of Tensors for Dynamic QoS Analysis," *In Proc. of the 2021 IEEE Int. Conf. on Web Services. (ICWS2021) (Regular)*, Chicago, IL, USA , 2021, pp. 560-568.

[8] X. Luo, Y. Zhou, Z. Liu, and M. Zhou, "Fast and Accurate Non-negative Latent Factor Analysis on High-dimensional and Sparse Matrices in Recommender Systems," *IEEE Transactions on Knowledge and Data Engineering*, DOI: 10.1109/TKDE.2021.3125252.

[9] D. Wu, and X. Luo, "Robust Latent Factor Analysis for Precise Representation of High-dimensional and Sparse Data", *IEEE/CAA Journal of Automatica Sinica*, vol. 8, no. 4, pp. 796-805, 2021.

[10] X. Luo, H. Wu, Z. Wang, J. Wang, and D. Meng, "A Novel Approach to Large-Scale Dynamically Weighted Directed Network Representation," *IEEE Transactions on Pattern Analysis and Machine Intelligence*, vol. 44, no. 12, pp. 9756-9773, 2022.

[11] M. Chen, C. He, and X. Luo. "MNL: A Highly-Efficient Model for Large-scale Dynamic Weighted Directed Network Representation," *IEEE Transactions on Big Data*, DOI: 10.1109/TBDATA.2022.3218064.

[12] H. Li, P. Wu, N. Zeng, Y. Liu, and F. Alsaadi, "A Survey on Parameter Identification, State Estimation and Data Analytics for Lateral Flow Immunoassay: from Systems Science Perspective," *International Journal of Systems Science*, DOI:10.1080/00207721.2022.2083262.

[13] D. Wu, X. Luo, M. Shang, Y. He, G. Wang, and X. Wu, "A Data-Characteristic-Aware Latent Factor Model for Web Services QoS Prediction," *IEEE Transactions on Knowledge and Data Engineering*, vol. 34, no. 6, pp. 2525-2538, 2022.

[14] Z. Liu, G. Yuan, and X. Luo, "Symmetry and Nonnegativity-Constrained Matrix Factorization for Community Detection," *IEEE/CAA Journal of Automatica Sinica*, vol. 9, no. 9, pp. 1691-1693, 2022.

[15] H. Wu, X. Luo, and M. C. Zhou. "Neural Latent Factorization of Tensors for Dynamically Weighted Directed Networks Analysis," *In Proc. of the 2021 IEEE Int. Conf. on Systems, Man, and Cybernetics*, Melbourne, Australia, 2021, pp. 3061-3066.

[16] X. Luo, Y. Yuan, S. Chen, N. Zeng, and Z. Wang, "Position-Transitional Particle Swarm Optimization-Incorporated Latent Factor Analysis," *IEEE Transactions on Knowledge and Data Engineering*, vol. 34, no. 8, pp. 3958-3970, 2022.

[17] J. Chen, X. Luo, and M. Zhou, "Hierarchical Particle Swarm Optimization-incorporated Latent Factor Analysis for Large-Scale Incomplete Matrices," *IEEE Transactions on Big Data*, vol. 8, no. 6, pp. 1524-1536, 2022.

[18] X. Luo, M. Zhou, S. Li, L. Hu, and M. Shang, "Non-negativity Constrained Missing Data Estimation for High-dimensional and Sparse Matrices from Industrial Applications," *IEEE Transactions on Cybernetics*, vol. 50, no. 5, pp. 1844-1855, 2020.

[19] W. Zhang, H. Sun, X. Liu, and X. Guo, "Temporal QoS-aware web service recommendation via non-negative tensor factorization," in *Proceedings of the 23rd international conference on World wide web*, New York, United States, 2014, pp. 585-596.

[20] X. Luo, H. Wu, M. Zhou and H. Yuan, "Temporal Pattern-aware QoS Prediction via Biased Non-negative Latent Factorization of Tensors," *IEEE Transactions on Cybernetics*, vol. 50 , no. 5, pp. 1798-1809, 2020.

[21] H. Wu, X. Luo, and M. Zhou, "Advancing Non-negative Latent Factorization of Tensors with Diversified Regularizations," *IEEE Transactions on Services Computing*, vol. 15, no. 3, pp. 1334-1344, 2022.

[22] X. Luo, M. Chen, H. Wu, Z. Liu, H. Yuan, and M. Zhou, "Adjusting Learning Depth in Non-negative Latent Factorization of Tensors for Accurately Modeling Temporal Patterns in Dynamic QoS Data," *IEEE Transactions on Automation Science and Engineering*, vol. 18, no. 4, pp. 2142-2155, 2022.

[23] Z. Liu, X. Luo, and M. Zhou, "Symmetry and Graph Bi-regularized Non-Negative Matrix Factorization for Precise Community Detection," *IEEE Transactions on Automation Science and Engineering*, DOI: 10.1109/TASE.2023.3240335.

[24] H. Wu, X. Luo, M. C. Zhou, M. J. Rawa, K. Sedraoui, and A. Albeshri. "A PID-Incorporated Latent Factorization of Tensors Approach to Dynamically Weighted Directed Network Analysis," *IEEE/CAA Journal of Automatica Sinica*, vol. 9, no. 3, pp.533-546, 2022.

[25] H. Wu, X. Luo, and M. C. Zhou. "Discovering Hidden Pattern in Large-scale Dynamically Weighted Directed Network via Latent Factorization of Tensors," *In Proc. of the 17th IEEE Int. Conf. on Automation Science and Engineerin*, Lyon, France, 2021, pp. 1533-1538.

[26] Y. Song, Z. Zhu, M. Li, G. Yang, and X. Luo, "Non-negative Latent Factor Analysis-Incorporated and Feature-Weighted Fuzzy Double c-Means Clustering for Incomplete Data," *IEEE Transactions on Fuzzy Systems*, vol. 30, no. 10, pp. 4165-4176, 2022.

[27] X. Luo, M. Zhou, S. Li, and M. Shang, "An Inherently Non-negative Latent Factor Model for High-dimensional and Sparse Matrices from Industrial Applications," *IEEE Transactions on Industrial Informatics*, vol. 14, no. 5, pp. 2011-2022, 2018.

[28] Z. S Lin, and H. Wu. "Dynamical Representation Learning for Ethereum Transaction Network via Non-negative Adaptive Latent Factorization of Tensors," *In Proc. of the 2021 Int. Conf. on Cyber-physical Social Intelligence*, Beijing, China, pp. 1-6, 2021.

[29] X. Luo, J. Sun, Z. Wang, S. Li, and M. Shang, "Symmetric and Non-negative Latent Factor Models for Undirected, High Dimensional and Sparse Networks in Industrial Applications," *IEEE Transactions on Industrial Informatics*, vol. 13, no. 6, pp. 3098-3107, 2017.

[30] W. Li, Q. He, X. Luo, and Z. Wang, "Assimilating Second-Order Information for Building Non-Negative Latent Factor Analysis-Based Recommenders," *IEEE Transactions on System Man Cybernetics: Systems*, vol. 52, no. 1, pp. 485-497, 2021.

[31] X. Luo, M. Zhou, Y. Xia, Q. Zhu, A. Ammari, and A. Alabdulwahab, "Generating Highly Accurate Predictions for Missing QoS-data via Aggregating Non-negative Latent Factor Models," *IEEE Transactions on Neural Networks and Learning Systems*, vol. 27, no. 3, pp. 579-592, 2016.

[32] Y. Xu, Z. Wu, J. Chanussot, and Z. Wei. "Hyperspectral images super-resolution via learning high-order coupled tensor ring representation," *IEEE transactions on neural networks and learning systems*, vol. 31, no. 11, pp. 4747-4760, 2020.

[33] Z. Long, C. Zhu, J. Liu, and Y. Liu. "Bayesian low rank tensor ring for image recovery," *IEEE Transactions on Image Processing*, vol. *30*, pp. 3568-3580, 2021.

[34] Q. Zhao, G. Zhou, S. Xie, L. Zhang, and A. Cichocki. "Tensor ring decomposition," *arXiv preprint arXiv:1606.05535*, 2016.

[35] W. He, Y. Chen, N. Yokoya, C. Li, and Q. Zhao. "Hyperspectral super-resolution via coupled tensor ring factorization," *Pattern Recognition*, 122, 108280, 2022.

[36] J. Yu, G. Zhou, W. Sun, and S. Xie. "Robust to rank selection: Low-rank sparse tensor-ring completion," *IEEE Transactions on Neural Networks and Learning Systems*, DOI: 10.1109/TNNLS.2021.3106654.

[37] H. Wu, Y. Xia, and X. Luo. "Proportional-Integral-Derivative-Incorporated Latent Factorization of Tensors for Large-Scale Dynamic Network Analysis," *In Proc. of the 2021 China Automation Congress*, Beijing, China, 2021, pp. 2980-2984.

[38] J. Fang, Z. Wang, W. Liu, S. Lauria, N. Zeng, C. Prieto, F. Sikstrom, and X. Liu, "A New Particle Swarm Optimization Algorithm for Outlier Detection: Industrial Data Clustering in Wire Arc Additive Manufacturing," *IEEE Transactions on Automation Science and Engineering*, DOI:10.1109/TASE.2022.3230080.

[39] X. Luo, Y. Zhou, Z. Liu, L. Hu, and M. Zhou, "Generalized Nesterov's Acceleration-incorporated, Non-negative and Adaptive Latent Factor Analysis," *IEEE Transactions on Services Computing*, vol. 15, no. 5, pp. 2809-2823, 2022.

[40] N. Zeng, P. Wu, Z. Wang, H. Li, W. Liu, X. Liu, "A small-sized object detection oriented multi-scale feature fusion approach with application to defect detection," *IEEE Transactions on Instrumentation and Measurement*, vol. 71, no. 35, PP. 7-14, 2022.

[41] Xing. Su, M. Zhang, Y. Liang, Z. Cai, L. Guo, and Z. Ding, "A tensor-based approach for the QoS evaluation in service-oriented environments," *IEEE Transactions on Network and Service Management*, vol. 18, no. 3, pp. 3843-3857, 2021.

[42] S. Wang, Y. Ma, B. Cheng, F. Yang, and R. N. Chang, "Multi-dimensional QoS prediction for service recommendations," *IEEE Transactions on Services Computing*, vol. 12, no.1, pp. 47-57, 2019.